\documentclass[letterpaper,10pt,conference]{ieeeconf}

\IEEEoverridecommandlockouts
\usepackage{amsmath}
\usepackage{amssymb}
\usepackage{array}
\usepackage{booktabs}
\usepackage{CJKutf8}
\usepackage{cite}
\usepackage{graphicx}
\usepackage{placeins}
\usepackage{tabularx}
\usepackage{xcolor}
\usepackage{url}
\usepackage[hidelinks]{hyperref}

\graphicspath{{figures/}}

\title{\LARGE\bfseries
UnderGrasp: Envelopment-Oriented Grasp Refinement for an Underactuated Gripper
}

\author{Zhiwen Song$^{1\dagger}$, Ruidi Liu$^{2\dagger}$, Guanglu Jia$^{1*}$, Zuowei Chen$^{1}$, Hao Wen$^{1}$ and Qing Shi$^{1}$
\thanks{$^{\dagger}$These authors contributed equally to this work.}%
\thanks{$^{*}$corresponding authors.}%
\thanks{$^{1}$Zhiwen Song, Guanglu Jia, Zuowei Chen, Hao Wen and Qing Shi
are with Beijing Institute of Technology, Beijing, China}%
\thanks{$^{2}$Ruidi Liu
is with Beijing University of Technology, Beijing, China}%
}

\begin{document}
\maketitle
\thispagestyle{empty}
\pagestyle{empty}

\begin{CJK*}{UTF8}{gbsn}

\begin{abstract}
Underactuated grippers can form enveloping grasps through passive compliance,
while the terminal pose and single width predicted by generic grasp generators
do not fully capture the approach--closure process. Conventional collision
filtering also struggles to distinguish harmful interference from acceptable
contact. 
We present UnderGrasp, a post-hoc grasp refinement and evaluation framework 
for an underactuated gripper that reuses grasp candidates and quality priors 
from a frozen generator.
The model jointly encodes the target, local
environment, support surface, and approach-path geometry to propose diverse
combinations of insertion depth, pre-grasp width, and target closing width
for each candidate. These actions are ranked by combining separately
predicted envelopment and interaction quality with upstream priors,
balancing grasp feasibility, envelopment, and scene disturbance rather
than rejecting all environmental contact. The back end is trained solely
on configuration and interaction labels generated by gripper-specific
simulations, without real-robot grasp labels. Real-robot experiments with an
underactuated gripper show that UnderGrasp improves the overall success rate (SR)
and the envelopment rate among successful grasps (ER) from 0.55/0.12 to
0.82/0.55 with AnyGrasp, and from
0.42/0.08 to 0.65/0.36 with GraspNet. In contrast, increasing the fixed
insertion depth to 40~mm improves ER to 0.35 but reduces SR to 0.52,
supporting the value of adaptive refinement. For trials involving target
or environmental contact during approach, the conditional success rates increase from 0.36
to 0.89 and from 0.25 to 0.86, respectively, demonstrating more reliable
execution under contact.
Project page: \url{https://dishenliu2004-hub.github.io/undergrasp-site/}.
\end{abstract}


\section{INTRODUCTION}

\begin{figure}[t]
  \centering
  \includegraphics[width=\columnwidth]{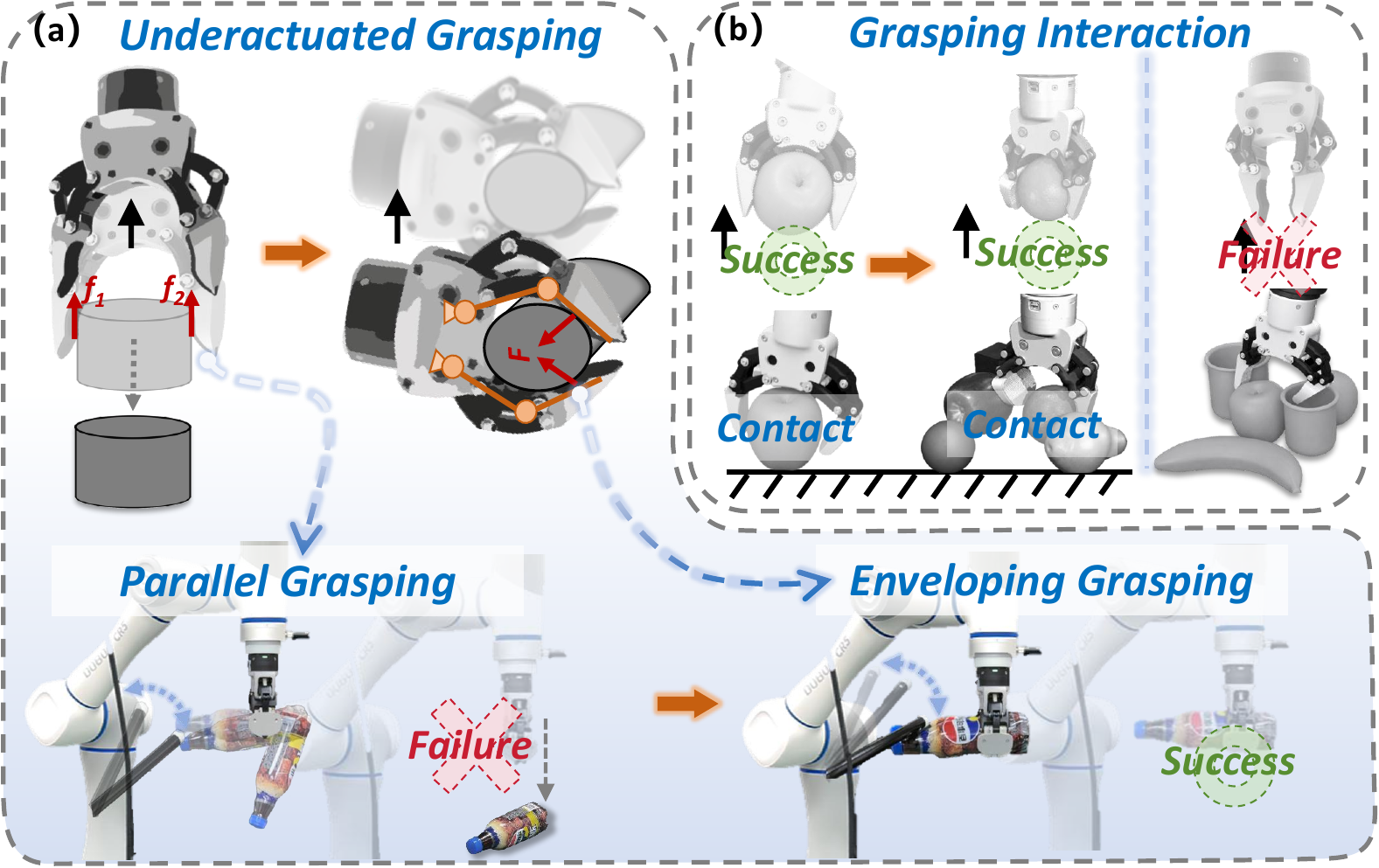}
  \caption{Motivation for underactuated grasp refinement. (a) Transition from a parallel grasp to 
  a robust enveloping grasp through passive adaptation. (b) 
  Interaction with the environment or target is not always harmful; moderate contact may improve grasping.}
  \label{fig:motivation}
\end{figure}

Stable and robust grasping of unknown objects in unstructured environments is
a key capability for general-purpose robotic manipulation~\cite{bicchi2000review,bohg2014survey}.
Underactuated grippers combine multi-link fingers with a single actuator 
and passive compliance, allowing them to conform to object geometry 
and achieve both parallel and enveloping grasps~\cite{dollar2010sdm,odhner2014compliant}. 
Enveloping grasps provide stronger geometric constraints and 
greater robustness to perception errors, execution deviations, 
and external disturbances~\cite{ko2024gravity},
as shown in Fig.~\ref{fig:motivation}(a).
Realizing these benefits requires grasp poses, contact geometry, and closing actions
that exploit the gripper's mechanical characteristics~\cite{roa2015quality,shao2020unigrasp}.
Large-scale grasp datasets and 3D deep networks have enabled 7-DoF
grasp generators, such as AnyGrasp and GraspNet~\cite{fang2023anygrasp,fang2020graspnet}, to efficiently generate and rank
candidates from RGB observations or point clouds while generalizing to unseen
objects and cluttered scenes. These methods are typically designed for
rigid parallel-jaw grippers and represent each grasp using a terminal end-effector
pose and a single gripper width, while the gripper's closing action must still 
be specified by the downstream execution policy.

Furthermore, existing methods often approximate grippers as combinations of simple 
geometric primitives for collision checking, limiting their ability to accurately 
represent the spatial occupancy of real multi-link grippers and its evolution during motion~\cite{fang2020graspnet,sundermeyer2021contact}.
Beyond this geometric limitation, collision-avoidance strategies for rigid parallel-jaw grippers 
typically regard environmental contact as undesirable interference~\cite{danielczuk2021object,yamada2026graspm}. For underactuated grippers, 
however, some contact is not necessarily detrimental, as illustrated in Fig.~\ref{fig:motivation}(b). 
Appropriate early contact with the environment or the target object may induce only minor disturbance 
while guiding the fingers toward a more favorable enveloping configuration, 
potentially improving subsequent grasp stability~\cite{eppner2015exploitation,bimbo2019edge}. Such contacts should therefore be evaluated 
by balancing the resulting gains in grasp quality against the associated disturbance costs.
Accordingly, post-processing of grasp candidates would benefit from considering not only 
the gripper's actual geometry but also the effects of contact on both grasp execution and 
the surrounding environment, thereby determining whether a contact is beneficial or acceptable.

To address these challenges, we propose 
a post-hoc grasp refinement and evaluation framework 
for an underactuated gripper (UnderGrasp), which reuses grasp candidates and their quality scores from 
a frozen grasp generator. 
It jointly encodes the target, local environment, support surface, and candidate-specific
approach-path point clouds to predict 
an insertion depth offset along the approach axis, a pre-grasp width offset, and 
a target closing width offset, 
yielding a refined stage-wise action from a generic grasp candidate. Ranking combines upstream grasp quality with predicted envelopment 
and interaction quality, balancing grasp feasibility, envelopment, and environmental 
contact costs and benefits. Trained solely on gripper-specific simulations,
the network uses geometry-focused supervision from enveloping configurations,
target capture, and scene changes to reduce reliance on precise physical parameter
identification. Real-robot experiments demonstrate transfer without real-world grasp labels.

The primary contributions are as follows:
\begin{itemize}
\item 
We introduce UnderGrasp, a post-hoc framework that reuses candidates and quality priors 
from a frozen grasp generator. 
It refines the generic grasp candidate into a stage-wise action,
adapting execution to target geometry and surrounding spatial constraints.

\item 
We propose a refinement-evaluation architecture. 
The refinement module generates local variants of each input grasp candidate, 
while the evaluation module predicts physical quantities used to 
evaluate their envelopment and interaction qualities. These results, 
together with the upstream grasp score, are used to rank the refined candidates 
based on grasp feasibility, envelopment quality, and the costs 
and benefits of environmental contact.

\item 
Real-robot experiments demonstrate that, with the back end trained solely 
in simulation and without real-robot grasp labels, the proposed method 
achieves effective enveloping grasps on diverse objects not used in back-end training while
better balancing grasp robustness and environmental interaction, 
thereby reducing harmful interference.

\end{itemize}

\section{RELATED WORK}

\subsection{7-DoF Grasp Generation}

Generic grasp pose generators typically represent parallel-jaw grasp 
candidates by their 3D position, orientation, and minimum opening width~\cite{sundermeyer2021contact,fang2020graspnet,fang2023anygrasp}. 
Contact-GraspNet generates dense candidates from observed contact points~\cite{sundermeyer2021contact}, 
GraspNet uses graspness to focus on high-quality regions~\cite{wang2021graspness}, and AnyGrasp 
further incorporates awareness of the center of gravity and cross-frame association 
to improve stability and dynamic robustness~\cite{fang2023anygrasp}. GraRe reranks candidates 
by combining local geometry and object context while keeping the detector 
frozen and the grasp poses and opening widths unchanged~\cite{yuan2026grare}. 
However, the predicted width characterizes the grasp candidate's geometry
but does not specify a complete closing command. For example, AnyGrasp's dynamic grasping
procedure invokes a separate gripper-closing command after reaching the selected pose~\cite{fang2023anygrasp}.
This representation alone does not specify how insertion depth, 
approach opening, and post-contact closing input should be coordinated 
to form an underactuated enveloping grasp.

\subsection{Gripper-Specific Grasping and Adaptation}
To bridge the gap between generic grasp generators and actual gripper mechanisms~\cite{shao2020unigrasp}, 
Parulekar et al. use conditional flow matching to adapt AnyGrasp candidates for 
a Fin-ray soft gripper~\cite{parulekar2026grasp}. AdapGrasp combines object stiffness and grasp affordances 
to predict initial and final states, aiming to reduce damage and improve stability~\cite{pu2026adapgrasp}. 
For underactuated grippers, Ko et al. use contact simulations to evaluate resistance 
to disturbances along the direction of gravity and guide enveloping grasp selection~\cite{ko2024gravity}. 
Our focus is the joint refinement of insertion depth, pre-grasp opening, and target
closing width for underactuated enveloping grasping, with separate supervision for
geometric envelopment and process-level interaction.

\subsection{Collision-Aware Grasping and Contact Interaction}
Environment-aware grasping methods typically use collision constraints
to improve execution feasibility. SceneCollisionNet predicts collisions
between query objects and the scene~\cite{danielczuk2021object}, while Grasp-MPC combines grasp value
with collision-avoidance and smoothness costs~\cite{yamada2026graspm}.
Environment-assisted grasping studies show that contact is not necessarily
detrimental: Eppner et~al. investigate the role of environmental constraints
in robust grasping~\cite{eppner2015exploitation}, while Bimbo et~al. exploit hand compliance and
environmental constraints for edge grasps~\cite{bimbo2019edge}.
However, collision filtering based on geometric overlap alone cannot
distinguish interference that displaces the target or prevents capture
from acceptable, low-disturbance contact. The above environment-assisted
methods focus on specific contact strategies, limiting their direct use
for comparing contact outcomes across grasp candidates.
For underactuated grippers, these approaches do not fully account for
contact-induced changes in finger configuration and scene disturbance
when distinguishing harmful interference, acceptable contact,
and contact that may facilitate envelopment.

\section{METHOD}

\subsection{Overview}

\begin{figure*}[t]
  \centering
  \includegraphics[width=0.97\textwidth]{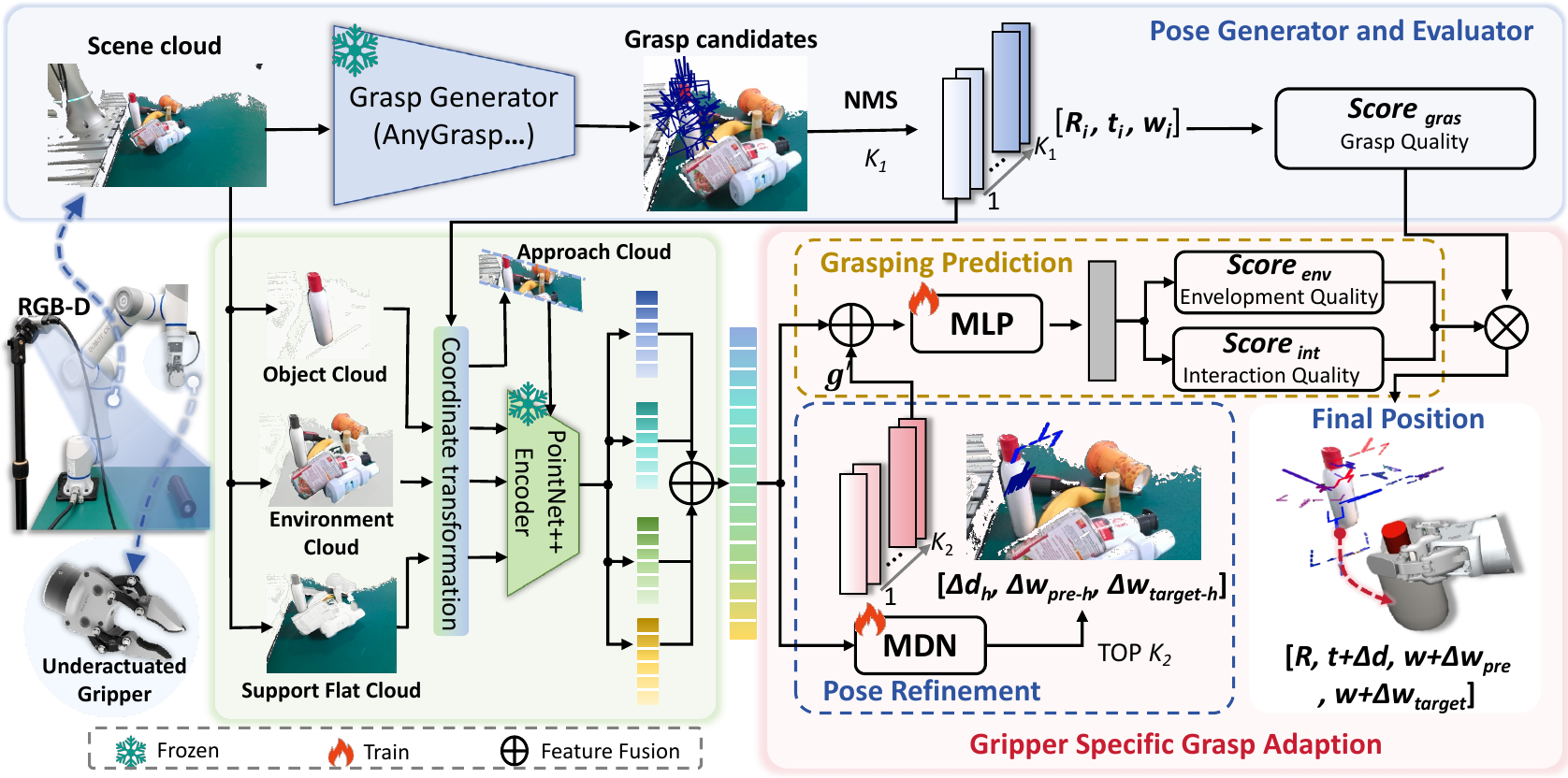}
  \caption{UnderGrasp pipeline. A frozen generator produces grasp candidates,
  which are filtered by NMS. Candidate-centered target, environment, support,
  and approach-path point clouds condition the action proposal and evaluation
  heads; their scores are combined with the generator score for final ranking.}
  \label{fig:pipeline}
\end{figure*}

UnderGrasp adapts the output of a generic grasp generator to an underactuated gripper
using four candidate-centered point-cloud inputs. As shown in
Fig.~\ref{fig:pipeline}, the frozen front end predicts grasp candidates
$\boldsymbol{g}_i=[\mathbf{R}_i,\mathbf{t}_i,W_i]$ with quality
$S_{\mathrm{G},i}$, where $\mathbf{R}_i\in SO(3)$, $\mathbf{t}_i\in\mathbb{R}^3$,
and $W_i$ denote the orientation, position, and nominal opening width, respectively~\cite{fang2020graspnet,fang2023anygrasp}.
Non-maximum suppression (NMS) in $SE(3)$ removes geometrically redundant poses and
retains $K_1$ candidates for subsequent adaptation.

For each candidate, we form four point clouds in its local frame: the target, nearby
environment, support surface, and scene points along the candidate's approach path.
PointNet++~\cite{qi2017pointnet} extracts a geometric feature from each cloud; the four
features are fused with the candidate parameters into $\mathbf{z}_i$. The first three inputs
describe the target and its surroundings, while the approach-path cloud explicitly
captures clearance and potential early contact along the incoming motion. Conditioned on
$\mathbf{z}_i$, the proposal head generates $K_2$ approach--closure actions.
For each proposed action, the evaluation network takes the refined grasp configuration
$\boldsymbol{g}'_{ih}$ together with the fused feature $\mathbf{z}_i$ as input and predicts
its envelopment and interaction outcomes.
The front end thus
provides scene-level grasp hypotheses, and the back end determines how each hypothesis
should be executed by the specified gripper.

\subsection{Adaptive Grasp Execution Refinement}
 
The nominal width $W_i$ does not specify the opening used during approach or the closing
command after contact. For an underactuated gripper, these two stages jointly determine
clearance and passive envelopment~\cite{dollar2010sdm,odhner2014compliant,ko2024gravity}. We therefore predict an insertion offset $\Delta d_{ih}$
and two width offsets $\Delta W_{\mathrm{pre},ih}$ and
$\Delta W_{\mathrm{target},ih}$ relative to the front-end width $W_i$. Specifically,
$W_{\mathrm{pre},ih}=W_i+\Delta W_{\mathrm{pre},ih}$ is the opening used during
approach, whereas $W_{\mathrm{target},ih}=W_i+\Delta W_{\mathrm{target},ih}$ is the
target closing width. The $h$-th action associated with candidate $i$ is
\begin{equation}
\begin{split}
  \boldsymbol{g}'_{ih}
  = \big[&\mathbf{R}_i,\ \mathbf{t}_i +\mathbf{e}_z \Delta d_{ih},\
          W_{\mathrm{pre},ih},\
          W_{\mathrm{target},ih}\big],
\end{split}
  \label{eq:refined-action}
\end{equation}
where $i\in\{1,\ldots,K_1\}$ indexes the retained candidates,
$h\in\{1,\ldots,K_2\}$ indexes the proposed actions, and
$\mathbf{e}_z$ is the unit approach direction. The rotation $\mathbf{R}_i$ and the position
components perpendicular to the approach direction remain unchanged; $\Delta d_{ih}$ translates the
terminal pose along $\mathbf{e}_z$, $W_{\mathrm{pre},ih}$ controls the approach
clearance, and $W_{\mathrm{target},ih}$ specifies the closure command. The two width
offsets independently adapt the approach and closing stages while retaining $W_i$ as the
front-end width prior. This parameterization
preserves the geometric grasp hypothesis and adds only the gripper-dependent variables
needed for its execution.

A single candidate may support several valid depth--width combinations. To avoid averaging
distinct execution modes, we employ a mixture density network (MDN)~\cite{bishop1994mdn}
to model $p_{\phi}(\mathbf{a}\mid\mathbf{z}_i)$, where
$\mathbf{a}=[\Delta d,\Delta W_{\mathrm{pre}},\Delta W_{\mathrm{target}}]$ is the
local offset vector
and $\phi$ denotes the proposal-network parameters. Each mixture component predicts its
probability, mean, and diagonal scale within the valid action range. The resulting
$K_2$ proposals retain diverse approach--closure strategies around the same front-end
candidate; their grasp quality is determined by the evaluation head rather than the MDN
likelihood.

\subsection{Envelopment and Interaction Quality}

The evaluation network separates geometric envelopment, defined as the extent to which
the fingers wrap around the target, from process-level interaction.
Its envelopment branch predicts the left and right envelopment values $E_L$ and $E_R$; the
approach branch predicts contact probabilities $c_e,c_t,c_s$, disturbances $D_e,D_t$,
approach-stage envelopment $E_{\mathrm{app}}$, and support force $F_s$; and the closure branch
predicts target-capture probability $c_g$ and over-closing risk $R_W$. The probability
outputs and continuous quantities are combined into envelopment quality
$S_{\mathrm{env}}$ and interaction quality $S_{\mathrm{int}}$ as follows.

\noindent\textbf{Envelopment Quality.}
Because an underactuated grasp may begin with unilateral contact but should favor
bilateral wrapping, we define
\begin{equation}
S_{\mathrm{env}}=
\begin{cases}
(1-a_{\mathrm{env}})H^{q_{\mathrm{env}}}
 +a_{\mathrm{env}}U^{q_{\mathrm{env}}}, & U>\varepsilon,\\
0, & U\leq\varepsilon
\end{cases},
\label{eq:envelopment-quality}
\end{equation}
where $E_L,E_R\in[0,1]$ are the normalized envelopment values of the left and
right distal links. At each active-joint position, the contact-free distal-link
angle defines the non-enveloping reference, and only additional inward angular
deviation from this reference is counted as envelopment. $U=\max(E_L,E_R)$
measures the stronger unilateral envelopment, and
$H=2E_LE_R/(E_L+E_R)$ measures bilateral consistency. We set $H=0$ when
$E_L+E_R\leq\varepsilon$, where $\varepsilon>0$ prevents division by zero.
$a_{\mathrm{env}}\in[0,1]$ balances the two terms, and $q_{\mathrm{env}}>0$ controls
their response to partial envelopment. Consequently, $U$ preserves useful one-sided
wrapping, whereas $H$ favors a balanced final configuration.

\noindent\textbf{Interaction Quality.}
We first aggregate contact events, scene disturbances, and excessive closure
into a measure of harmful interaction risk
\begin{equation}
R_{\mathrm{bad}}
=\lambda_e c_e+\lambda_t c_t
+\mu_e D_e+\mu_t D_t
+\lambda_W R_W,
\label{eq:harmful-risk}
\end{equation}
where $c_e$ and $c_t$ denote the probabilities of contacting the surrounding
environment and prematurely contacting the target during approach,
respectively. Here, the approach stage extends from the pre-grasp pose to the start
of commanded gripper closure. $D_e,D_t\in[0,1]$ are the corresponding disturbance magnitudes,
and $R_W\in[0,1]$ measures the over-closing risk. The non-negative coefficients
$\lambda_e$ and $\lambda_t$ independently weight the two contact events,
$\mu_e$ and $\mu_t$ weight the associated environment and target disturbances,
and $\lambda_W$ controls the contribution of excessive closure. This
formulation allows the occurrence and severity of different interaction risks
to be calibrated independently while accumulating concurrent hazards.

Support contact is not always harmful~\cite{eppner2015exploitation,bimbo2019edge}. 
Gentle contact may increase the contact area between the gripper and the object, 
thereby improving grasp robustness, whereas ineffective or excessive contact should be suppressed. 
We express this effect by
\begin{equation}
\begin{split}
G_s=\exp\!\Big\{c_s\kappa_s\big[&E_{\mathrm{app}}q_F(F_s)
 -(1-E_{\mathrm{app}})\\
&-(1-q_F(F_s))\big]\Big\},
\end{split}
\label{eq:support-gain}
\end{equation}
where $c_s\in[0,1]$ is the support-contact probability,
$E_{\mathrm{app}}\in[0,1]$ measures envelopment formed during approach, and $F_s\geq0$ is
the normal support force. The force quality is
$q_F(F)=2^{-(F/F_{50})^2}$, with $F_{50}>0$ denoting the calibrated half-quality force;
$\kappa_s>0$ controls the overall support contribution. The first term in brackets
rewards envelopment obtained with low force, while the remaining terms penalize support
contact that yields little envelopment or excessive force. Thus, $G_s$ is neutral when
$c_s=0$, larger than one for beneficial support, and smaller than one for harmful support.

The complete interaction quality is gated by the probability of capturing the target:
\begin{equation}
S_{\mathrm{int}}=c_g^{\eta}\exp(-R_{\mathrm{bad}})G_s,
\label{eq:interaction-quality}
\end{equation}
where $c_g\in[0,1]$ is the target-capture probability and $\eta>0$ controls its influence.
$\exp(-R_{\mathrm{bad}})$ suppresses harmful interaction, while $G_s$ rewards or penalizes
support contact according to its outcome. The capture term prevents support interaction
alone from producing a high score.

\noindent\textbf{Joint Ranking.}
Finally, the gripper-specific scores are combined with the front-end grasp prior:
\begin{equation}
S_{\mathrm{final},ih}
= \bar{S}_{\mathrm{G},i}^{\alpha}
  \left(1+\beta S_{\mathrm{env},ih}\right)
  \left(S_{\mathrm{int},ih}\right)^{\gamma},
\label{eq:joint-ranking}
\end{equation}
where $\bar S_{\mathrm{G},i}$ is the calibrated score of candidate $i$, and
$S_{\mathrm{env},ih}$ and $S_{\mathrm{int},ih}$ evaluate its $h$-th action. Calibration
places scores produced by different front ends on a comparable scale~\cite{guo2017calibration}.
$\alpha$, $\beta$, and $\gamma$ control the grasp prior, envelopment gain, and interaction quality, 
respectively, and can be adjusted to emphasize different selection criteria for different requirements.
The factor $1+\beta S_{\mathrm{env},ih}$ retains the original candidate prior while allowing
envelopment to improve its ranking.
For all experiments, we set $\beta=2$, $\lambda_W=0.5$, and $\kappa_s=0.2$;
the same values are used for different front-end grasp generators.

\subsection{Simulation Supervision and Learning}

\begin{figure}[t]
  \centering
  \includegraphics[width=0.96\columnwidth]{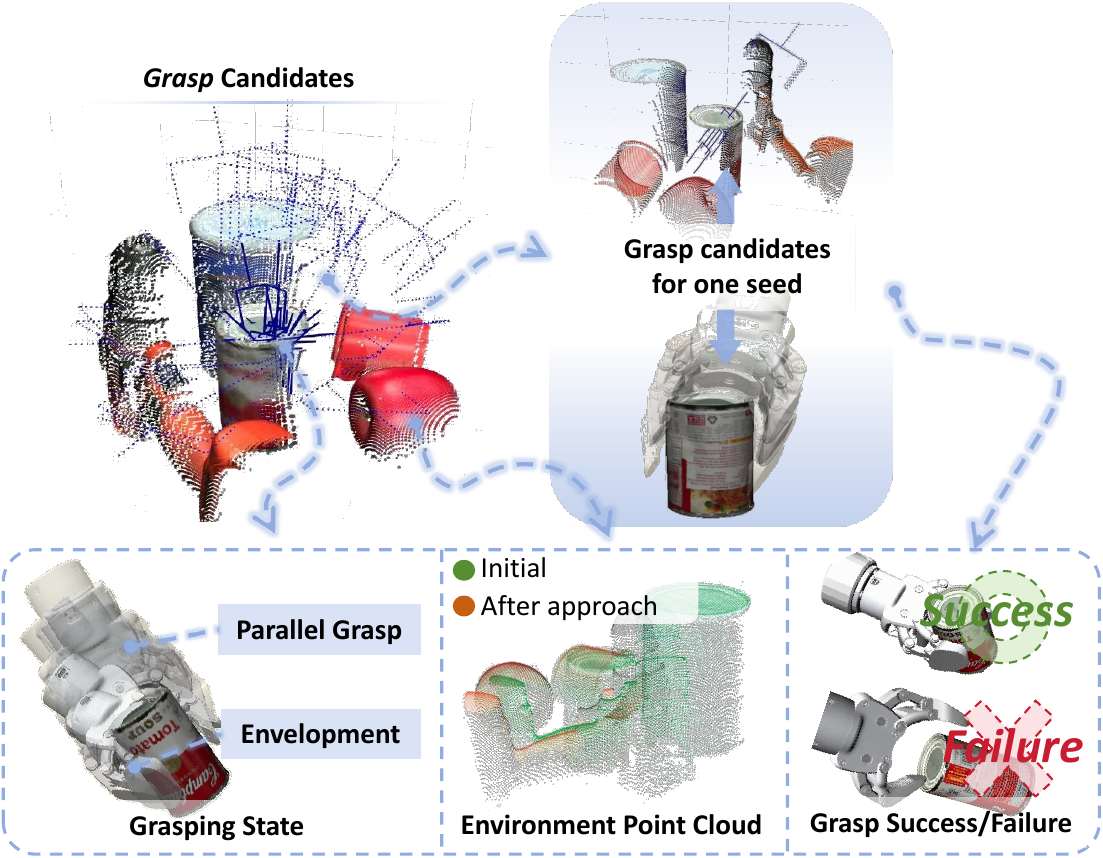}
  \caption{Simulation data generation and supervision. For each seed grasp
  candidate, multiple approach--closure actions are executed to record finger
  configurations, approach-induced scene changes, and terminal grasp outcomes.}
  \label{fig:simulation-supervision}
\end{figure}

The outcome variables above require intermediate contact and closure states that are
difficult to annotate on a real robot~\cite{mousavian2019graspnet,ko2024gravity}. We therefore execute the local actions in
MuJoCo~\cite{todorov2012mujoco}, as illustrated in
Fig.~\ref{fig:simulation-supervision}. Approach trajectories provide supervision for
$c_e,c_t,c_s,D_e,D_t,E_{\mathrm{app}}$, and $F_s$; the terminal state provides
$E_L,E_R,c_g$, and $R_W$. Contact and capture events use binary training labels, while
their network outputs are probabilities during inference. Envelopment, disturbance, force,
and closure risk are learned as continuous quantities.

The over-closing label compares the commanded closure with the state actually reached by
the active joints:
\begin{equation}
R_W=\operatorname{clip}\!\left(
\frac{f_\theta(W_{\mathrm{target}})-\theta_{\mathrm{end}}
-\delta_\theta}{\tau_\theta},0,1\right).
\label{eq:over-closing-risk}
\end{equation}
Here, $f_\theta$ is the calibrated mapping from target width $W_{\mathrm{target}}$ to
active-joint angle, $\theta_{\mathrm{end}}$ is the terminal joint angle,
$\delta_\theta\geq0$ is the admissible mismatch, and $\tau_\theta>0$ is the
normalization scale. $\operatorname{clip}(\cdot,0,1)$ bounds the label to $[0,1]$.
Accordingly, $R_W=0$ when the commanded closure is reached within tolerance and increases
with the unattained closing command. Equation~\eqref{eq:over-closing-risk} is used only
for simulation supervision; the closure branch predicts $R_W$ directly at inference.

We train the probability outputs $c_g,c_e,c_t,c_s$ using binary cross-entropy and the
continuous outputs $E_L,E_R,D_e,D_t,E_{\mathrm{app}},F_s,R_W$ using Smooth L1 loss~\cite{girshick2015fast}.
The proposal MDN is optimized by quality-weighted likelihood, assigning greater weight to
actions with higher envelopment and interaction scores; a repulsion term maintains diversity
among mixture components. Training proceeds from the evaluation branches to the proposal
head, followed by joint fine-tuning. The generic grasp generator remains frozen, and
simulation is not required during deployment.

Training data are generated by executing simulated grasps in clean and cluttered scenes.
The clean scenes comprise three configurations for each of 12 objects, while the cluttered
scenes comprise four configurations for each of four representative target objects selected
from the same set of 12 objects. For each configuration,
we retain $K_1=40$ seed grasp candidates after non-maximum suppression. For each candidate,
we uniformly sample eight insertion offsets $\Delta d$ within $0$--$50~\mathrm{mm}$ and
set the target closing width $W_{\mathrm{target}}$ to $15\%$, $25\%$, $45\%$, or $70\%$
of the original width $W$, yielding $K_2=32$ approach--closure actions. The pre-grasp width offset
$\Delta W_{\mathrm{pre}}$ is randomly sampled from $[-5\%,15\%]$ of $W$.
Each action starts from a pre-grasp pose retracted by $10~\mathrm{cm}$ along the approach
axis and executes the approach and closure stages in simulation to obtain the corresponding
supervision labels. This procedure yields 46,080 action samples from clean scenes and
20,480 from cluttered scenes, resulting in 66,560 training samples in total.

\section{EXPERIMENTS}
\subsection{Experimental Setup}

\noindent\textbf{Platforms.}
We evaluate UnderGrasp in MuJoCo and on a real-robot platform. Both platforms use the
same underactuated gripper model and the stage-wise approach and closure procedure
described in Sec.~III-B. The physical system consists of a Dobot CR5 6-axis robot arm,
an OmniPicker two-finger underactuated gripper, and a fixed external Intel RealSense
D435i RGB-D camera. The observed point cloud is transformed into the robot base frame
using calibrated camera-to-robot extrinsics. The simulated gripper matches its physical
counterpart in geometry, joint limits, and control sequence.

\begin{figure}[!htbp]
  \centering
  \includegraphics[width=0.82\columnwidth]{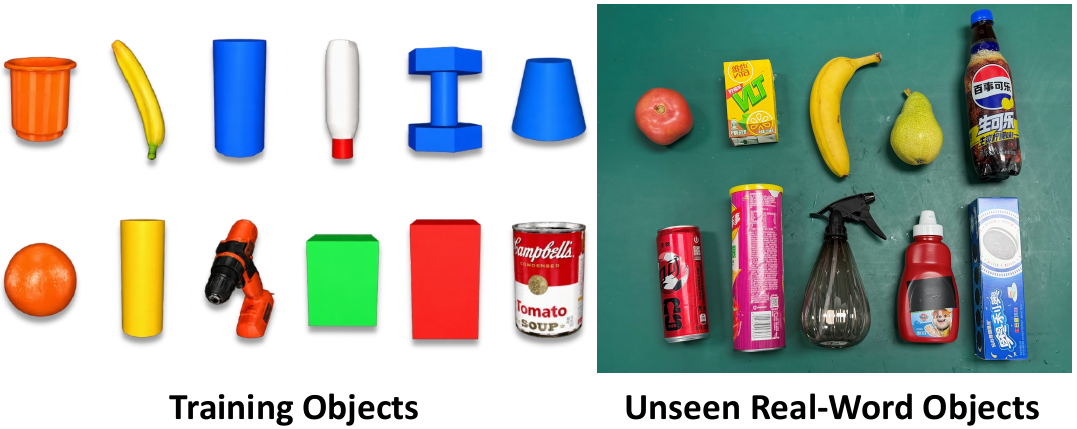}
  \caption{Objects used in the experiments. The left and right panels show
  the objects used for simulation training and real-robot evaluation,
  respectively.}
  \label{fig:experimental-objects}
\end{figure}

\noindent\textbf{Objects and Scenarios.}
The training objects and real-robot test objects are shown in
Fig.~\ref{fig:experimental-objects}. Twelve objects covering diverse geometries,
sizes, and masses are used for training, while 10 objects excluded from UnderGrasp back-end
training are used for real-robot evaluation. Experiments are conducted in both clean and cluttered scenes,
with object poses randomized in both settings. The experiments are organized into
four groups: refined grasp pose evaluation, real-robot grasping performance,
motion robustness, and environmental interaction.

\noindent\textbf{Evaluation Protocol.}
Target detection and segmentation are performed using YOLO-World~\cite{cheng2024yoloworld} and SAM~\cite{kirillov2023segment}, respectively.
All methods use the same RGB-D observations, robot configuration, motion controller,
and candidate budget. For the real-robot grasping evaluation, each object is tested
in 6 trials, including 3 in clean scenes and 3 in cluttered scenes, with
the scene reset after every attempt. Since generic grasp generators do not specify closing commands,
we set the baselines' target closing width to 10\% of the gripper's maximum opening,
as preliminary real-robot tests favored smaller closing targets within the tested range.
Network parameters are selected using the training set.

\subsection{Grasping Refinement}

\begin{figure*}[t]
  \centering
  \includegraphics[width=0.97\textwidth]{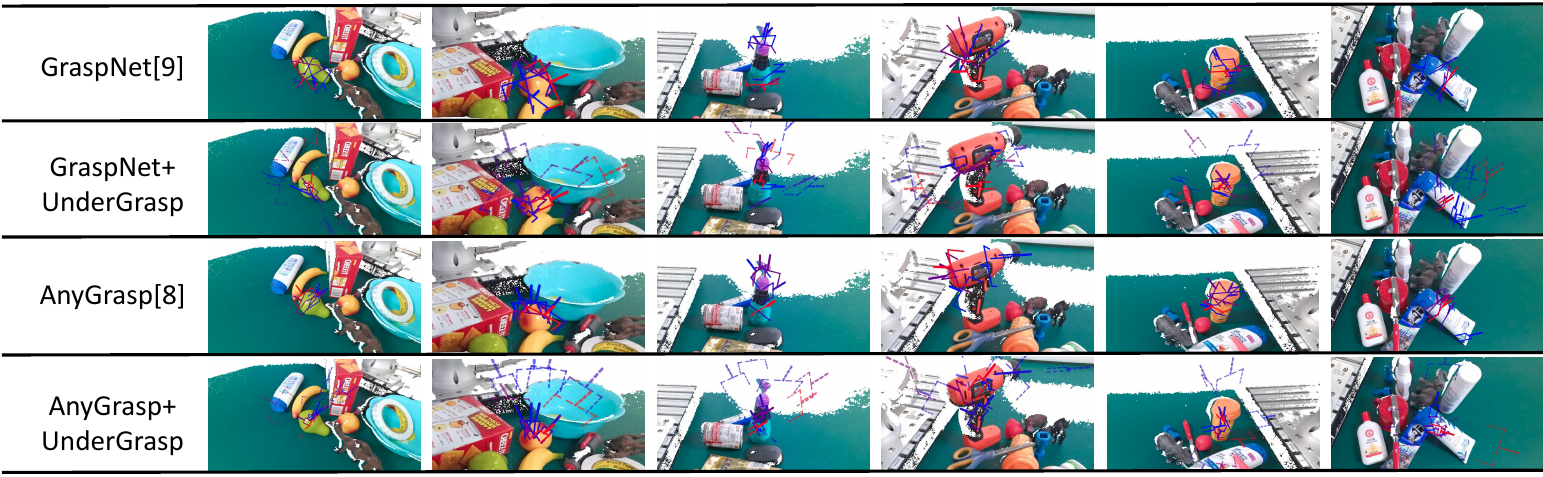}
  \caption{Qualitative grasp proposals in six cluttered GraspNet-1Billion scenes.
  From top to bottom, the rows show GraspNet, GraspNet + UnderGrasp, AnyGrasp,
  and AnyGrasp + UnderGrasp. Each panel displays the five highest-ranked
  candidates, with the top candidate in red; solid and dashed outlines mark
  target and pre-grasp configurations, respectively.}
  \label{fig:refinement-cloud}
\end{figure*}

To examine how UnderGrasp modifies generic parallel-jaw grasp hypotheses, we use
GraspNet~\cite{fang2020graspnet} and AnyGrasp~\cite{fang2023anygrasp} as two representative
front ends. Candidate grasps are generated for objects selected from six cluttered scenes
in GraspNet-1Billion. Fig.~\ref{fig:refinement-cloud} shows the five highest-ranked
candidates produced by the two original generators and their variants refined by UnderGrasp.
The best candidate in each scene is highlighted in red; solid and dashed gripper outlines
denote the target and pre-grasp configurations, respectively. The four configurations
select noticeably different actions. After refinement, the highest-ranked actions tend
to be perpendicular to the object's major axis while preserving unobstructed fingertip paths,
which provides a more favorable enveloping configuration.
This change in the orientations of the highest-ranked grasps results from reranking
the original candidates; the rotation of each individual candidate remains unchanged.

\begin{figure}[!htbp]
  \centering
  \includegraphics[width=0.84\columnwidth]{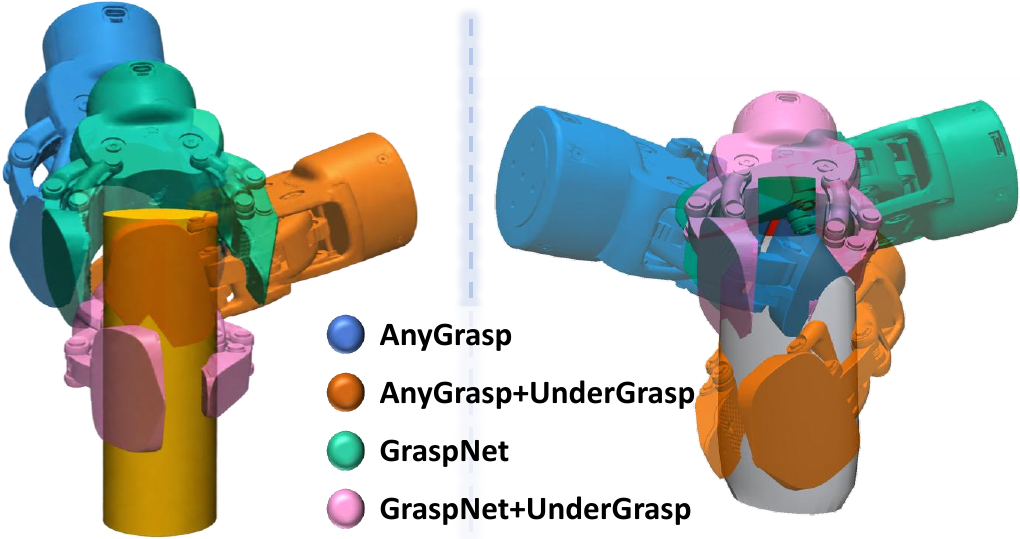}
  \caption{Simulated terminal gripper configurations for two representative
  objects. Colored overlays compare AnyGrasp and GraspNet executions before and
  after UnderGrasp refinement under the same object-specific initial conditions.}
  \label{fig:simulated-closure}
\end{figure}

To compare the resulting executions directly, we further apply the highest-ranked action
from each configuration to two representative objects in simulation. As shown in
Fig.~\ref{fig:simulated-closure}, the actions refined by UnderGrasp produce terminal finger
configurations that more closely conform to the object surface, indicating a stronger
tendency to form enveloping grasps than the directly transferred grasps.

\subsection{Grasping Performance and Robustness}

\noindent\textbf{Overall Performance.}
To evaluate whether UnderGrasp can refine generic 7-DoF grasp candidates into
effective actions for an underactuated gripper, Table~\ref{tab:grasp-performance}
compares five configurations using AnyGrasp and GraspNet as front ends.
Because the OmniPicker does not close like the ideal parallel-jaw gripper assumed
by these generators, a 10-mm insertion offset along the approach axis is added
to both unadapted baselines. AnyGrasp and GraspNet serve as base generators for
evaluating direct transfer, while AnyGrasp + fixed depth uses a 40-mm insertion
offset to test whether the improvement achieved by UnderGrasp can be attributed simply
to deeper insertion. All methods use the same robot configuration. The 10 test objects are randomly
placed, and each is evaluated in 6 trials, including 3 in clean scenes
and 3 in cluttered scenes, resulting in 60 trials per method. A trial is
successful if the robot grasps the target, returns to its initial pose, and
maintains the grasp for 3~s without dropping it. Success rate (SR) is the
fraction of successful trials, while envelopment rate (ER) is the fraction of
successful grasps that form an enveloping grasp.
A successful grasp is classified as enveloping when contact between the inner finger
segments and the object is followed by additional inward rotation of the distal joints,
causing the fingers to wrap around the object.

\begin{table*}[t]
  \caption{Real-robot grasp success rate (SR) and envelopment rate (ER, measured among successful grasps).}
  \label{tab:grasp-performance}
  \centering
  \footnotesize
  \setlength{\tabcolsep}{3.5pt}
  \renewcommand{\arraystretch}{1.15}
  \begin{tabularx}{\textwidth}{@{}
      >{\raggedright\arraybackslash}p{3.5cm}
      *{6}{>{\centering\arraybackslash}X}@{}}
    \toprule
    Method
      & \multicolumn{2}{c}{Single object}
      & \multicolumn{2}{c}{Cluttered scene}
      & \multicolumn{2}{c}{Overall} \\
    \cmidrule(lr){2-3}\cmidrule(lr){4-5}\cmidrule(lr){6-7}
      & SR $\uparrow$ & ER $\uparrow$
      & SR $\uparrow$ & ER $\uparrow$
      & SR $\uparrow$ & ER $\uparrow$ \\
    \midrule
    GraspNet~\cite{fang2020graspnet}
      & 0.53 & 0.13
      & 0.30 & 0.00
      & 0.42 & 0.08 \\
    AnyGrasp~\cite{fang2023anygrasp}
      & 0.63 & 0.11
      & 0.47 & 0.14
      & 0.55 & 0.12 \\
    AnyGrasp + 40-mm insertion
      & 0.57 & 0.47
      & 0.47 & 0.21
      & 0.52 & 0.35 \\
    GraspNet + UnderGrasp
      & 0.77 & 0.35
      & 0.53 & 0.38
      & 0.65 & 0.36 \\
    AnyGrasp + UnderGrasp
      & \textbf{0.83} & \textbf{0.60}
      & \textbf{0.80} & \textbf{0.50}
      & \textbf{0.82} & \textbf{0.55} \\
    \bottomrule
  \end{tabularx}
\end{table*}

Increasing the fixed insertion depth from 10 to 40~mm raises ER from 0.12 to
0.35, but reduces SR from 0.55 to 0.52. This shows that deeper insertion alone
does not guarantee more reliable grasping: the resulting contacts also depend
on object geometry, gripper pose, and the interaction process. A fixed depth
may therefore push differently shaped objects away or cause unfavorable
collisions before a stable grasp is formed.

In contrast, UnderGrasp adaptively refines the execution parameters according
to the candidate grasp and local geometry, while jointly considering the
generator score, envelopment quality, and interaction quality. It achieves
0.82 SR and 0.55 ER with AnyGrasp. With GraspNet as the front end, SR and ER
also improve from 0.42/0.08 to 0.65/0.36, indicating that the gains are not
specific to one grasp generator. We further observe that the relatively low success rates of the baselines 
are partly caused by the non-ideal parallel closing behavior of the underactuated gripper: 
some objects are squeezed out of the grasp region during closure, 
while others slip or are ejected as the robot returns to its initial pose. 
In addition, given the limited number of trials, a single failure can noticeably affect 
the reported success rate.

\begin{figure}[!htbp]
  \centering
  \includegraphics[width=\columnwidth]{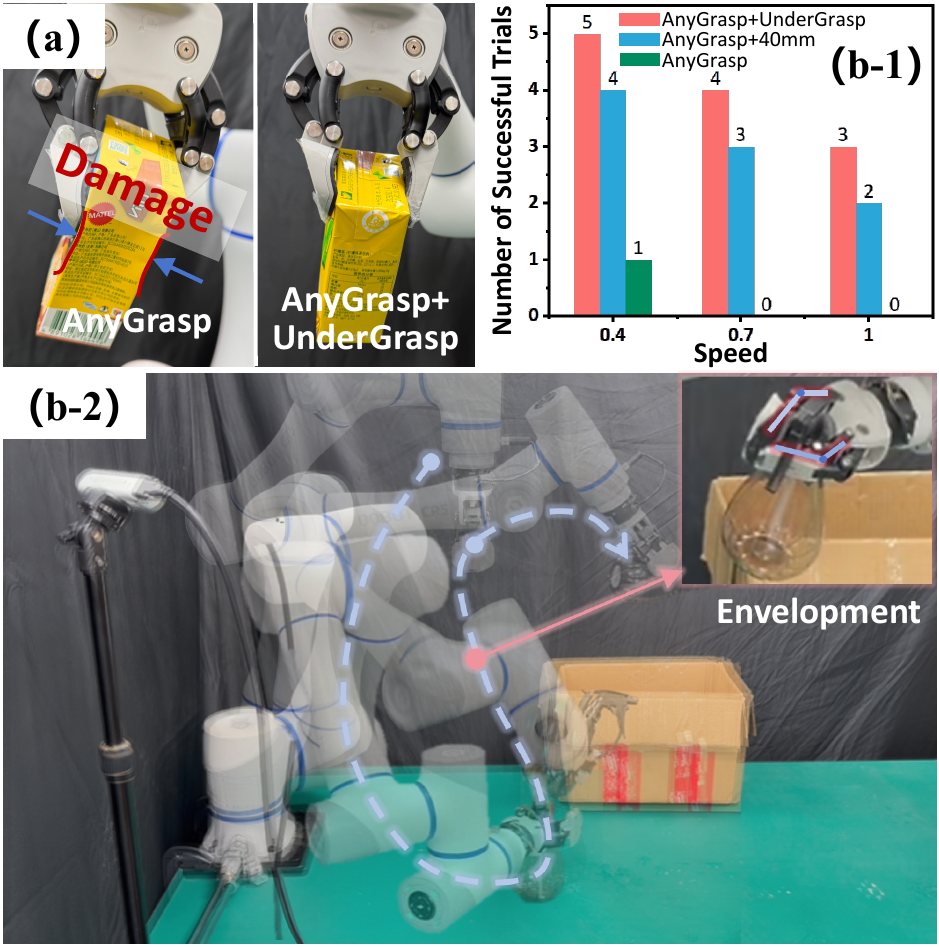}
  \caption{Grasp robustness. (a) Excessive closure damages the object, whereas
  a suitable target width avoids damage. (b-1) Number of successful post-grasp
  transports at three normalized robot speeds. (b-2) Representative real-robot
  transport sequence.}
  \label{fig:robustness}
\end{figure}

\noindent\textbf{Grasp Robustness.}
Furthermore, we evaluate the closure safety of UnderGrasp and 
its post-grasp robustness under different motion conditions.
Fig.~\ref{fig:robustness}(a) shows that the baseline, 
which applies a fixed closing width across different grasp poses, 
can over-close the gripper and damage the object. In contrast, 
UnderGrasp predicts the target closing width and incorporates a soft constraint on
excessive closure through an over-closing risk penalty in action ranking,
allowing the closing action to adapt to each grasp pose and thereby reducing object damage.

In the motion robustness experiment, we compare AnyGrasp, AnyGrasp + fixed depth, and AnyGrasp + UnderGrasp. 
The grasped object is a watering can filled with 500~mL of water. Only trials with a successful initial 
grasp are included in the robustness evaluation. All initial grasps evaluated for
AnyGrasp + UnderGrasp are enveloping grasps. After grasping, the robot lifts the object,
returns to its initial pose, and then transports it to a designated placement location. 
The robot speed is set to 40\%, 70\%, and 100\% of its maximum speed, with five trials conducted 
at each setting. The experimental procedure and results are shown in Fig.~\ref{fig:robustness}(b). 
AnyGrasp + UnderGrasp performs slightly better than the fixed-depth baseline and substantially 
better than the original AnyGrasp. Together with the observed enveloping configurations,
these results support the effectiveness of UnderGrasp for robust post-grasp transport
under the tested motion conditions.

\subsection{Environmental Interaction}

\begin{figure}[!htbp]
  \centering
  \includegraphics[width=0.84\columnwidth]{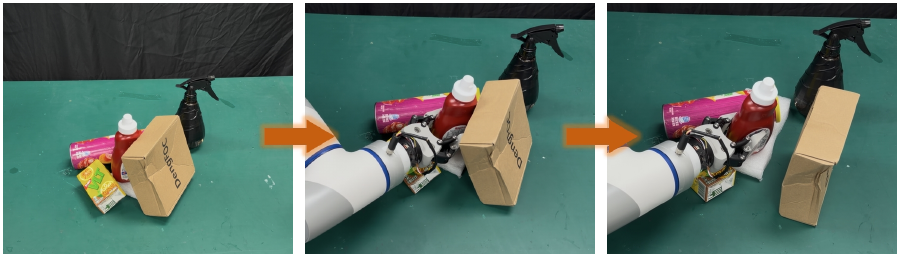}
  \caption{Real-robot grasp with environmental contact in clutter. During
  the approach to the target bottle, the gripper displaces a neighboring box but
  still achieves a robust grasp.}
  \label{fig:environment-interaction}
\end{figure}

We further evaluate how UnderGrasp balances environmental interaction and grasp success. 
Fig.~\ref{fig:environment-interaction} shows that the grasp actions generated by UnderGrasp 
can form configurations that are more favorable for stable grasping after contacting the environment, 
while introducing only limited environmental disturbance.

We further analyze contact events in the same real-robot grasping trials reported in
Table~\ref{tab:grasp-performance} for AnyGrasp and AnyGrasp + UnderGrasp, comprising
60 trials per method on 10 objects across clean and cluttered scenes.
Table~\ref{tab:interaction-strategies} summarizes grasp success conditioned on target
or environmental contact during approach. Here, target contact and environmental contact refer to
contacts occurring during the approach from the pre-grasp pose, before the gripper
begins its commanded closure. Target-contact SR and Env.-contact SR denote the grasp
success rates conditioned on target contact and environmental contact, respectively.
Each conditional rate is computed over trials with the corresponding contact, while
Overall SR denotes the success rate over all grasping trials. AnyGrasp has 11 trials
with target contact and 8 with environmental contact, whereas AnyGrasp + UnderGrasp
has 19 and 14, respectively. Despite
involving more contact, UnderGrasp achieves substantially higher conditional success rates, 
suggesting that contacts produced by the baseline are often unfavorable, whereas UnderGrasp 
more effectively favors contacts that support grasping or introduce only limited disturbance 
by jointly considering grasp quality, envelopment, and interaction effects.

\begin{table}[!htbp]
  \caption{Grasp success conditioned on target or environmental contact during approach, with overall SR for comparison.}
  \label{tab:interaction-strategies}
  \centering
  \footnotesize
  \setlength{\tabcolsep}{4pt}
  \renewcommand{\arraystretch}{1.18}
  \begin{tabularx}{\columnwidth}{@{}
      l *{3}{>{\centering\arraybackslash}X}@{}}
    \toprule
    Method & \shortstack{Target-contact\\SR}
      & \shortstack{Env.-contact\\SR}
      & \shortstack{Overall\\SR} \\
    \midrule
    AnyGrasp~\cite{fang2023anygrasp}
      & 0.36 & 0.25 & 0.55 \\
    AnyGrasp + UnderGrasp
      & \textbf{0.89} & \textbf{0.86} & \textbf{0.82} \\
    \bottomrule
  \end{tabularx}
\end{table}

\section{CONCLUSION}

This paper presents UnderGrasp, a post-hoc adaptation method that converts generic 
grasp candidates into underactuated approach and closure actions. It refines insertion 
depth and opening widths based on the candidate grasp poses, then ranks the actions 
using envelopment quality, interaction quality, and the frozen generator score.
With AnyGrasp, real-robot SR increases from 0.55 to 0.82, while ER among successful
grasps increases from 0.12 to 0.55. With GraspNet, the corresponding rates rise
from 0.42/0.08 to 0.65/0.36. In trials involving target or environmental
contact during approach, the conditional success rates increase from 0.36/0.25 to 0.89/0.86
after refinement. These results support gripper-specific action refinement across
front ends and show improved completion despite contact.

Favorable enveloping poses do not always succeed on the real robot. Closing can
push a large, lightweight object away before the fingers surround it, and some
support contacts still cause failure. These cases may reflect discrepancies in
mass, friction, and contact response between simulation and reality. Future work
will use real execution data to correct simulation-derived supervision and expand
the training set to a wider range of materials, masses, and deformable objects,
particularly lightweight compliant objects.



\end{CJK*}

\begin{thebibliography}{99}
\interlinepenalty=10000

\bibitem{bicchi2000review}
A. Bicchi and V. Kumar,
``Robotic grasping and contact: A review,''
in \emph{Proc. IEEE Int. Conf. Robotics and Automation (ICRA)}, vol. 1,
2000, pp. 348--353, doi: 10.1109/ROBOT.2000.844081.

\bibitem{bohg2014survey}
J. Bohg, A. Morales, T. Asfour, and D. Kragic,
``Data-driven grasp synthesis---A survey,''
\emph{IEEE Transactions on Robotics}, vol. 30, no. 2, pp. 289--309,
2014, doi: 10.1109/TRO.2013.2289018.

\bibitem{dollar2010sdm}
A. M. Dollar and R. D. Howe,
``The highly adaptive SDM Hand: Design and performance evaluation,''
\emph{The International Journal of Robotics Research}, vol. 29, no. 5,
pp. 585--597, 2010, doi: 10.1177/0278364909360852.

\bibitem{odhner2014compliant}
L. U. Odhner \emph{et al.},
``A compliant, underactuated hand for robust manipulation,''
\emph{The International Journal of Robotics Research}, vol. 33, no. 5,
pp. 736--752, 2014, doi: 10.1177/0278364913514466.

\bibitem{ko2024gravity}
T. Ko, T. Ikeda, T. Stewart, R. Lee, and K. Nishiwaki,
``Gravity-aware grasp generation with implicit grasp mode selection for
underactuated hands,'' in \emph{Proc. IEEE/RSJ Int. Conf. Intelligent Robots
and Systems (IROS)}, 2024, pp. 2964--2970,
doi: 10.1109/IROS58592.2024.10801600.

\bibitem{roa2015quality}
M. A. Roa and R. Su\'arez,
``Grasp quality measures: Review and performance,''
\emph{Autonomous Robots}, vol. 38, pp. 65--88, 2015,
doi: 10.1007/s10514-014-9402-3.

\bibitem{shao2020unigrasp}
L. Shao, F. Ferreira, M. Jorda, V. Nambiar, J. Luo, E. Solowjow,
J. A. Ojea, O. Khatib, and J. Bohg,
``UniGrasp: Learning a unified model to grasp with multifingered robotic
hands,'' \emph{IEEE Robotics and Automation Letters}, vol. 5, no. 2,
pp. 2286--2293, 2020, doi: 10.1109/LRA.2020.2969946.

\bibitem{fang2023anygrasp}
H.-S. Fang, C. Wang, H. Fang, M. Gou, J. Liu, H. Yan, W. Liu, Y. Xie,
and C. Lu,
``AnyGrasp: Robust and efficient grasp perception in spatial and temporal
domains,'' \emph{IEEE Transactions on Robotics}, vol. 39, no. 5,
pp. 3929--3945, 2023, doi: 10.1109/TRO.2023.3281153.

\bibitem{fang2020graspnet}
H.-S. Fang, C. Wang, M. Gou, and C. Lu,
``GraspNet-1Billion: A large-scale benchmark for general object grasping,''
in \emph{Proc. IEEE/CVF Conf. Computer Vision and Pattern Recognition (CVPR)},
2020, pp. 11444--11453.

\bibitem{sundermeyer2021contact}
M. Sundermeyer, A. Mousavian, R. Triebel, and D. Fox,
``Contact-GraspNet: Efficient 6-DoF grasp generation in cluttered scenes,''
in \emph{Proc. IEEE Int. Conf. Robotics and Automation (ICRA)}, 2021,
pp. 13438--13444, doi: 10.1109/ICRA48506.2021.9561877.

\bibitem{danielczuk2021object}
M. Danielczuk, A. Mousavian, C. Eppner, and D. Fox,
``Object rearrangement using learned implicit collision functions,''
in \emph{Proc. IEEE Int. Conf. Robotics and Automation (ICRA)}, 2021,
pp. 6010--6017.

\bibitem{yamada2026graspm}
J. Yamada, A. Murali, A. Mandlekar, C. Eppner, I. Posner,
and B. Sundaralingam,
``Grasp-MPC: Closed-loop visual grasping via value-guided model predictive
control,'' in \emph{Proc. IEEE Int. Conf. Robotics and Automation (ICRA)},
2026.

\bibitem{eppner2015exploitation}
C. Eppner, R. Deimel, J. \'Alvarez-Ruiz, M. Maertens, and O. Brock,
``Exploitation of environmental constraints in human and robotic grasping,''
\emph{The International Journal of Robotics Research}, vol. 34, no. 7,
pp. 1021--1038, 2015, doi: 10.1177/0278364914559753.

\bibitem{bimbo2019edge}
J. Bimbo, E. Turco, M. Ghazaei Ardakani, M. Pozzi, G. Salvietti, V. Bo,
M. Malvezzi, and D. Prattichizzo,
``Exploiting robot hand compliance and environmental constraints for edge
grasps,'' \emph{Frontiers in Robotics and AI}, vol. 6, Art. no. 135, 2019,
doi: 10.3389/frobt.2019.00135.

\bibitem{wang2021graspness}
C. Wang, H.-S. Fang, M. Gou, H. Fang, J. Gao, and C. Lu,
``Graspness discovery in clutters for fast and accurate grasp detection,''
in \emph{Proc. IEEE/CVF Int. Conf. Computer Vision (ICCV)}, 2021,
pp. 15964--15973.

\bibitem{yuan2026grare}
J. Yuan, Y. Zhao, Y. Lv, C. Xu, S. Li, C. Deng, and S. Chen,
``GraRe: Grasp candidate re-ranking for frozen 6-DoF grasp detectors,''
\emph{arXiv preprint arXiv:2608.00946}, 2026.

\bibitem{parulekar2026grasp}
T. Parulekar, G. Shi, J. Pinskier, D. Howard, and J. J. Chung,
``Grasp synthesis matching from rigid to soft robot grippers using
conditional flow matching,'' \emph{arXiv preprint arXiv:2602.17110}, 2026.

\bibitem{pu2026adapgrasp}
M. Pu, C. Han, Z. Chai, T. Zhao, D. Wu, P. Wen, X. Ke, H. Ding,
and Z. Wu,
``AdapGrasp: A stiffness and grasp affordance dataset with a
transformer-based adaptive grasp model,'' in \emph{Proc. IEEE Int. Conf.
Robotics and Automation (ICRA)}, 2026.

\bibitem{qi2017pointnet}
C. R. Qi, L. Yi, H. Su, and L. J. Guibas,
``PointNet++: Deep hierarchical feature learning on point sets in a metric
space,'' in \emph{Advances in Neural Information Processing Systems},
vol. 30, 2017.

\bibitem{bishop1994mdn}
C. M. Bishop, ``Mixture density networks,'' Neural Computing Research Group,
Aston University, Birmingham, U.K., Tech. Rep. NCRG/94/004, 1994.

\bibitem{guo2017calibration}
C. Guo, G. Pleiss, Y. Sun, and K. Q. Weinberger,
``On calibration of modern neural networks,''
in \emph{Proc. Int. Conf. Machine Learning (ICML)},
Proceedings of Machine Learning Research, vol. 70, pp. 1321--1330, 2017.

\bibitem{mousavian2019graspnet}
A. Mousavian, C. Eppner, and D. Fox,
``6-DOF GraspNet: Variational grasp generation for object manipulation,''
in \emph{Proc. IEEE/CVF Int. Conf. Computer Vision (ICCV)}, 2019,
pp. 2901--2910.

\bibitem{todorov2012mujoco}
E. Todorov, T. Erez, and Y. Tassa,
``MuJoCo: A physics engine for model-based control,''
in \emph{Proc. IEEE/RSJ Int. Conf. Intelligent Robots and Systems (IROS)},
2012, pp. 5026--5033, doi: 10.1109/IROS.2012.6386109.

\bibitem{girshick2015fast}
R. Girshick,
``Fast R-CNN,'' in \emph{Proc. IEEE Int. Conf. Computer Vision (ICCV)},
2015, pp. 1440--1448.

\bibitem{cheng2024yoloworld}
T. Cheng, L. Song, Y. Ge, W. Liu, X. Wang, and Y. Shan,
``YOLO-World: Real-time open-vocabulary object detection,''
in \emph{Proc. IEEE/CVF Conf. Computer Vision and Pattern Recognition (CVPR)},
2024, pp. 16901--16911.

\bibitem{kirillov2023segment}
A. Kirillov, E. Mintun, N. Ravi, H. Mao, C. Rolland, L. Gustafson,
T. Xiao, S. Whitehead, A. C. Berg, W.-Y. Lo, P. Doll\'ar, and R. Girshick,
``Segment Anything,'' in \emph{Proc. IEEE/CVF Int. Conf. Computer Vision
(ICCV)}, 2023, pp. 4015--4026.

\end{thebibliography}
\end{document}